\def\year{2019}\relax
\documentclass[letterpaper]{article} 
\usepackage{aaai19}  
\usepackage{times}  
\usepackage{helvet} 
\usepackage{courier}  
\usepackage[hyphens]{url}  
\usepackage{graphicx} 
\usepackage{graphicx}  
\usepackage{todonotes}
\usepackage{booktabs}       
\usepackage{tabularx}       

\title{Multimodal Dataset of Human-Robot Hugging Interaction}

\nocopyright  

\author{Kunal Bagewadi, Joseph Campbell, and Heni Ben Amor \\ 
School of Computing, Informatics, and Decision Systems Engineering\\ 
Arizona State University\\
Tempe, Arizona 85281\\
kbagewad@asu.edu 
}

\begin{document}

\maketitle
\begin{abstract}
A hug is a tight embrace and an expression of warmth, sympathy and camaraderie.
Despite the fact that a hug often only takes a few seconds, it is filled with details and nuances and is a highly complex process of coordination between two agents.
For human-robot collaborative tasks, it is necessary for humans to develop trust and see the robot as a partner to perform a given task together.
Datasets representing agent-agent interaction are scarce and, if available, of limited quality.
To study the underlying phenomena and variations in a hug between a person and a robot, we deployed Baxter humanoid robot and wearable sensors on persons to record \textit{353} episodes of hugging activity.
\textit{33} people were given minimal instructions to hug the humanoid robot for as natural hugging interaction as possible.
In the paper, we present our methodology and analysis of the collected dataset.
The use of this dataset is to implement machine learning methods for the humanoid robot to learn to anticipate and react to the movements of a person approaching for a hug.
In this regard, we show the significance of the dataset by highlighting certain features in our dataset.
\end{abstract}

\section{Introduction}
\label{intro}

Collaborative robots that live and work alongside human partners could radically transform a variety of application domains from manufacturing to health care, or the services sector. 
Instead of relying on repeated commands by a human operator, such robots autonomously blend their behavior with that of a human interaction partner, e.g. lend a hand to a co-worker lifting a heavy object. 
However, for this vision to become an everyday-reality, a theoretical foundation is needed that allows for the specification of collaborative interactions between humans and robots.

In this research task, we performed a human subject study to collect a dedicated set of motion tracking data of hugging interaction between humans and a robot. 
The dataset is used to better understand the inherent challenges and provide a training set for subsequent machine learning.
For the dataset, we enrolled \textit{33} people to get variation in the hug activity.
These are mostly graduate students in the age group of 25 to 35 years (27 male and 6 female participants).
No personal identifying data was collected.
From these participants, we recorded a total of \textit{353} hug episodes. 
Each hug episode consisted of a person walking straight towards the robot, hugging the robot and walking back to the starting position. 
We used Baxter research robot for the experiment because it is a humanoid robot and has arms that resemble human arms.
The duration of the hug, the style of body touching, the pressure of the body, and the activity of arms and hands disclose the intensity of relationship. 
The more frequent the hugs, the closer is the relationship \cite{hug_meaning}.
To capture and quantify these aspects, we mounted sensors on the person and Baxter, the humanoid robot, to collect data of hugging interaction.
We plan to release the dataset in form of csv files soon on our lab's website \textit{https://interactive-robotics.engineering.asu.edu/}.

\begin{figure*}[ht]     
\centering
\vspace*{-15mm}
\includegraphics[width=\textwidth, scale=0.13]{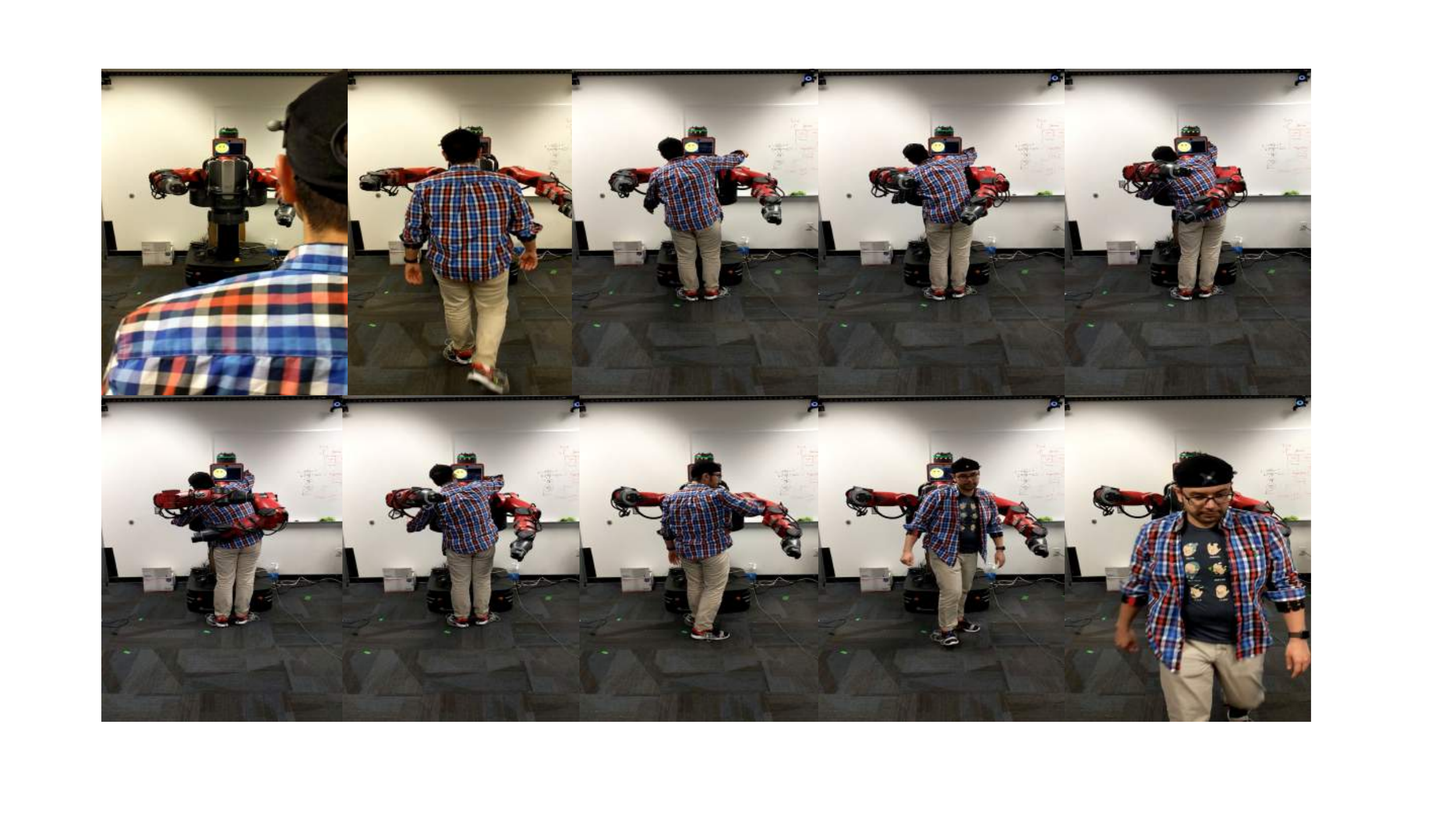}
\vspace*{-15mm}
\caption{A hug episode from POV of person's starting position. The person walks towards the robot, hugs it and walks back.
The manually-controlled robot opens its arms, hugs the person and opens the arms again to end the hug.}
\label{fig:hug_screenshots}
\end{figure*}

\section{Motivation}
\label{motivation}

In the traditional robot programming paradigm, a human engineer is required to foresee all important interaction parameters and implement control routines that generate appropriate robot responses. 
Unfortunately, even for moderately complex interaction scenarios, this approach becomes intractable. 
This is particularly true for physical and continuous interaction scenarios that are not based on turn-taking between the partners. 
The deployment of intelligent systems in homes and work places requires machine learning methods that go beyond the traditional supervised-learning approach. 
New methods are needed that combine both human intention recognition, as well as control of cyber-physical-systems, e.g. robots. 
The development of these methods could enable the next-generation of intelligent machines to adapt to the specific user or customer at hand. 
The result is a substantial improvement in quality-of-life; in particular for people who may have a disability or other circumstance that differentiates them from the average user. 
Especially given the growing population of senior citizens, it is critical to devise the next-generation of machines that safely engage in physical touch and interaction with a human user/partner.
The ability to bring humans and machines together in a safe, symbiotic relationship also enables a number of new application domains such as elderly care, mental-wellness therapy \cite{companion_robot}, \cite{dog_robot}, \cite{cat_robot}.
Experiments on psychological behavior reveal that robotic touching and hugging have a positive influence on the human partner and that humans tend to be proactive in HRI (human-robot interaction) \cite{robot_touch}, \cite{prosocial}.
Anthropomorphic robot appearance increases trust in the human partner in HRI \cite{robot_trust}.

The most pure form of physical contact and interaction in humans is “hugging”.
In humans, the repeated hugging of friends and family allows us to learn to pick up on even the smallest social cues and adapt our movements to the person being hugged.
To develop a robot using learning and interaction algorithms, we need a feature-rich dataset of the activity that we expect the robot to learn \cite{ML}. 
The key aim of our research so far has been to collect a multimodal dataset from the two agents, human and robot, involved in hugging activity.
Robot hugging can be used in introduction or greeting stage of any HRI task (e.g., assembly line), social robotics (e.g., therapy robot) or even for entertainment (e.g., amusement park).


\section{Related Work}
\label{rel_work}

Based on human-human hugging experiments, recent work in \cite{hug_model} develops a hug request model to find that humans utter a greeting as soon as they approach other human and that humans begin hand motion for a hug up to 0.4s after approaching other human. 
They implement the model on human-robot hugging experiment to find similar timings and that voice greeting and hand-motion have an effect on human emotions.
In our experiment, the humanoid robot began opening its arms to indicate hugging intention when the human started approaching for hug.
Researchers in \cite{involuntary_robot} developed a two-wheeled robot for natural looking, involuntary motion of the robot during activities like hugging, handshake.
Experiments in \cite{playful_robot} investigate full-body gesture recognition using inertial sensors on a small 37cm humanoid robot for playful interactions between humans and the robot.
They used SVM with 77\% average accuracy to classify 13 full-body gestures like stand up, lay down, back-n-forth, hug.

There have been studies that use toy-sized robot for activities that include hugging.
Work in \cite{hug_toy} describes design and implementation of affective feedback using projected avatars of their robot. They also presented user feedback through a questionnaire.
\cite{hug_teddy}, \cite{hug_platform_teddy} describe the design and effect of teddy-bear shaped robots on young patients.
For development of a toy, researchers in \cite{child_hug} measured the maximum force resulting in 2.263 psi exerted by children on a doll during a hug.
\cite{cat_robot} presents a cat-robot for interaction, that includes hugging, with children with autism.

There is research that analyzes social effects and social settings that encourage a hug from humans.
For instance, \cite{soft_warm} shows that humans prefer soft and warm robot hugs.
Our study does not focus on encouraging a hug but rather to collect and analyze feature-rich hugging dataset to use for robot learning. 
There is no dataset or experiments of hugging between a full-sized humanoid robot and a human that can be used for learning the hug behavior for the robot.


\section{Experiment Setup}
\label{exp_setup}

\begin{figure}[ht]
\centering
\includegraphics[scale=0.45]{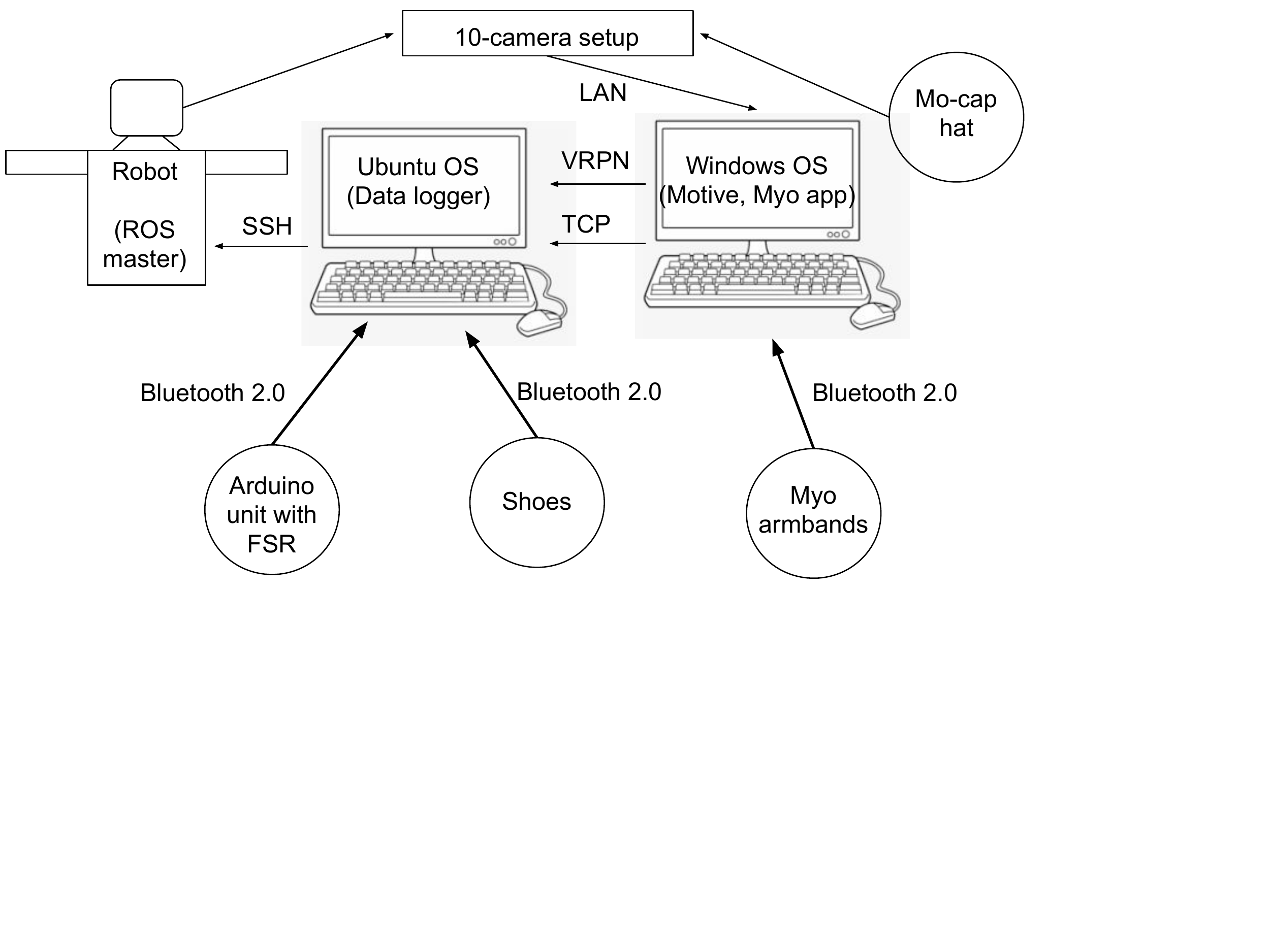}
\vspace{-35mm}
\caption{Summary of hardware and communication protocols used in experiment setup. The circular shapes denote sensors. The Ubuntu computer is used for data logging and controlling the robot.}
\label{fig:setup_diagram}
\end{figure}

The software in the experiment is based on Robot Operating System (ROS) framework. 
A computer executes nodes that collect data from sensors and also acts as data-logger that records data published on sensor topics and robot topics to save in a rosbag format.
OptiTrack's 10-camera setup in the room and Motive software tracks the reflective markers attached on the person's hat and robot's head at 120Hz.
The mo-cap(motion-capture) setup and Myo armband has Windows OS support only.
So we used a Windows computer to collect that data and transmit it to data-logger computer via Virtual-Reality Peripheral Network(VRPN) and TCP connections respectively.
A brief description of the robot and the sensors shown in Figure \ref{fig:setup_diagram} is given below.

\subsection{Robot}
\begin{figure}[ht]
\centering
\includegraphics[scale=0.08]{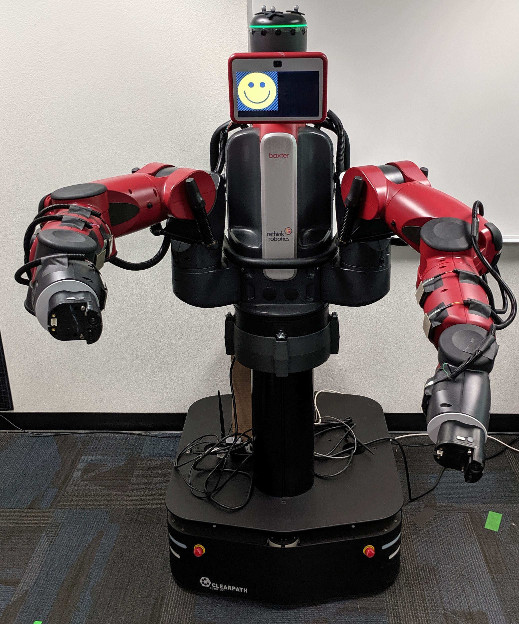}
\caption{Baxter robot in initial pose. Three motion capture(mo-cap) markers on its head.}
\label{fig:baxter_init}
\end{figure}

The hugging action requires the robot to be as human-like and safe as possible for close physical interaction with humans.
Baxter is a humanoid, anthropomorphic robot with \textit{two arms}, each with \textit{seven} degree-of-freedom(DOF)\cite{Baxter_specs}.
It has elastic actuators at the joints which provide a hard, spring like motion when the robot arms are pushed against an object; a person in this case.
So it is safe to operate the robot's arms for hugging, without injuring the person.
Each joint can be controlled at programmable speed. 
The robot is controlled by keyboard and joint speed is set to \textit{0.2rad/s} for safe operation.
The Baxter robot's ROS firmware publishes the angles at 100Hz.
Figure \ref{fig:baxter_init} shows the robot in the initial pose that indicates a request for hug.
The robot can be seen mounted on a mobile base but the robot's base was not moved during the experiment.

\begin{figure}[ht]
\centering
\includegraphics[scale=0.45]{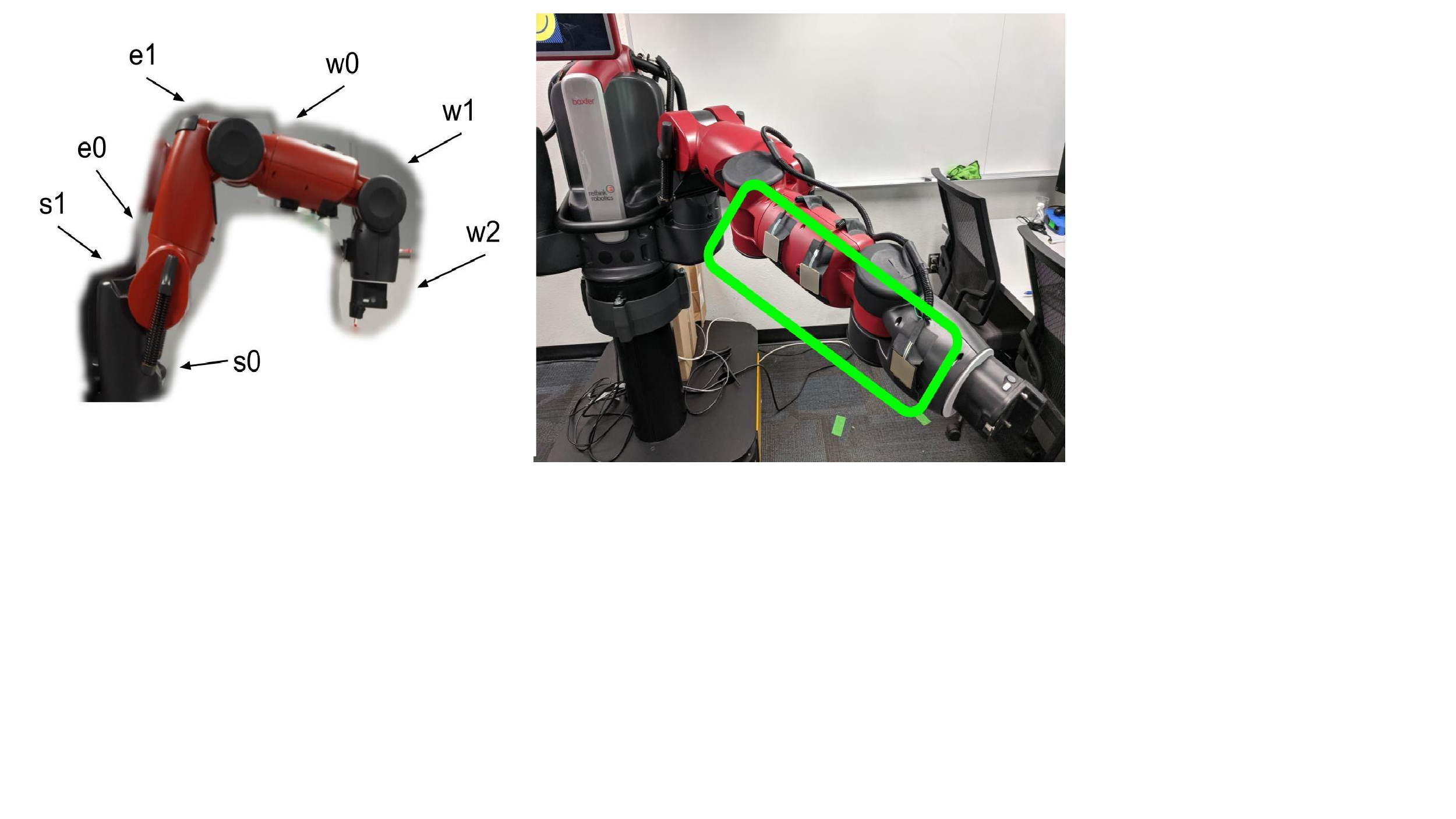}
\vspace{-30mm}
\caption{Left: Labels of all \textit{seven} DOF of one arm. \textit{s} =shoulder, \textit{e} =elbow, \textit{w} =wrist. 
Right: \textit{Three} FSRs highlighted on robot's left arm}
\label{fig:joint_labels}
\end{figure}

For sensing the force applied by the robot on the person's body during the hug, we used Ohmite FSR01CE force-sensitive resistors (FSR).
These sensors do not provide with exact force amount in Newtons but are sensitive enough to distinguish no-touch, soft and hard touch.
Three FSRs are mounted on each arm of the robot at upper forearm, lower forearm and wrist, as seen in Figure \ref{fig:joint_labels}.
Empirically, it was found that when a person hugs the Baxter robot, the person touches at least one of these locations of the robot's arms.
The FSRs are read by an Arduino unit which sends out FSR data to the data-logging computer via Bluetooth.

\subsection{Wearable sensors}

We used wearable sensors for persons hugging the robot to capture their position and various actions during the hug activity.
These light weight, wearable sensors are convenient and do not interfere with the person's movement. 
The sensors worn by the participants were Myo armbands, pressure sensing shoes and a hat with motion capture(mo-cap) markers.

Myo armbands\cite{Myo_specs} are worn on both forearms. 
They have accelerometer, gyroscope and EMG sensors for recording acceleration, orientation and muscle activity in the arms of the person. 
 
A hat populated with markers is worn by the person in the experiment. 
The position of the hat is tracked by the mo-cap setup, so that we have the position and orientation of the person in the room. 
The robot is also tracked by using markers on its head, as seen in Figure \ref{fig:baxter_init}.
The Myo armband and mo-cap data is read by a Windows computer which sends that data to the data-logging computer in real-time using TCP socket connection and VRPN respectively.

\begin{figure}[ht]
\centering
\hspace*{5mm}
\includegraphics[scale=0.5]{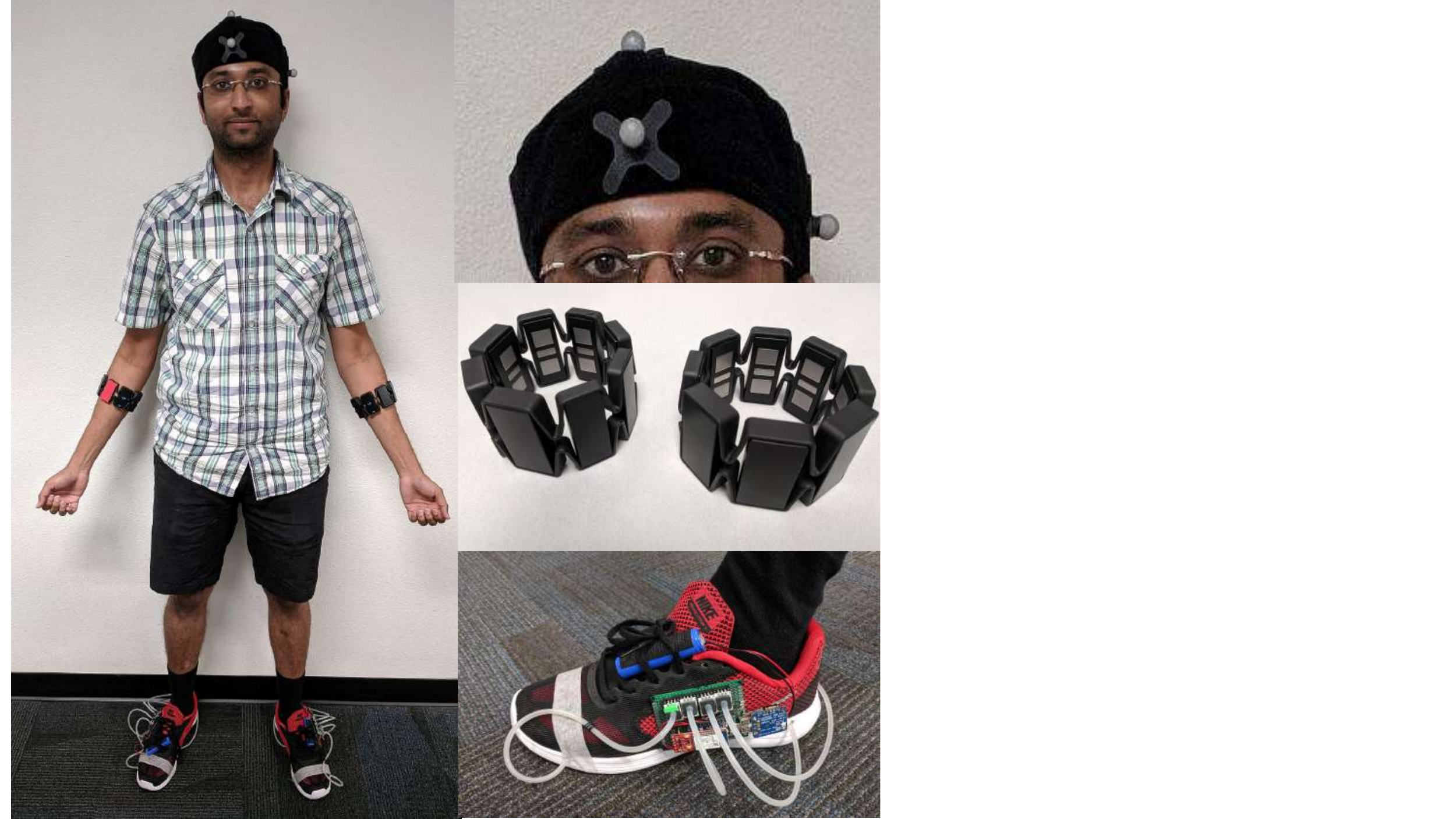}
\caption{Right: Enlarged images of the sensors: mo-cap hat, Myo armbands, pressure-sensing shoes. 
Left: These sensors worn by a person}
\label{fig:sensors_person}
\end{figure}

We rigged a regular pair of sport shoes with barometric pressure sensors at heel, toe, inner metatarsel and outer metatarsel locations of the foot in the shoe. 
An Arduino unit stitched on each shoe reads these sensors and sends the data using Bluetooth to the data-logging computer.
Figure \ref{fig:sensors_person} shows all these sensors worn by a person.

\subsection{Procedure}
\label{procedure}

The experiment was carried out in a closed room with 10-camera, motion-capture setup.
The participant was briefed about the experiment and was told only the following for instructions: "Walk towards the robot and hug the robot. The robot will be manually controlled to hug you. After the hug, walk back to the starting position".
The instructor sat behind the participant to supervise the experiment, controlled the robot with a keyboard and operated the data-logging computer.
The starting position of the participant was 8 feet away from robot, directly facing the robot.
We collected at least 10 hugs samples from each person and each hug episode lasted for approximately 30\textit{s}. 
In total, we have 353 hug samples from the experiment.
Figure \ref{fig:hug_screenshots} shows video grabs of a hug episode.

\section{Features}
\label{features}

The two categories of features of the dataset are the robot features and human features. 
The features originating from the robot are robot's joint angles and FSR values.
These form the robot features.
The human features are extracted from the output of sensors worn by the person. 
Table \ref{tab:feature_summary} summarizes the features which are explained in detail below.

\subsection{Robot}
All joint angles of the robot were recorded during the course of the experiment.
Since the robot has 7-DOF on each arm, the 14 joint angles form 14 features that capture the robot state during the experiment.
However roll type of DOF available for upper arm and lower arm was hardly ever used.

The FSRs on both of the robot's arms are read by 10-bit ADC on the Arduino unit.
The ADC's output is proportional to the amount of force exerted by the robot's arm on the person while hugging.
The FSR values form 6 features that give information about where the person touched the robot's arm while hugging.
It also tells us how tightly to grasp the person while hugging.
For reference, we saw that no touch gives ADC output of nearly zero while a hard touch yields ADC output of around 700, as seen in Figure \ref{fig:fsr_values}.
The reflective markers on the robot's head record the position and orientation of the robot in the room.

\subsection{Person}

The Myo armbands on the person's forearms measure acceleration and orientation of the arms. 
We wanted that information in order to capture the movement of the arms since they are used for hugging.
The data from the shoes is proportional to the pressure exerted at four locations of each foot of the person.
This feature is useful in identifying whether the person is walking or stationary.
Walking state can be for approaching the robot for a hug or walking away after the hug.
Stationary state can be for standing away from the robot, like the starting position of the experiment, or standing very close to the robot for the duration of the hug.
The mo-cap hat has markers which are tracked by the mo-cap setup.
Since the hat is worn by the person, we have the person's position and orientation in the room.

\begin{table}[ht]
\centering
\begin{tabular}{@{}lll@{}}
\toprule
\textbf{Device}                                              & \textbf{Features}                                                                           & \textbf{Hypothesis}                                                                            \\ \midrule
Robot                                                        & Joint angles                                                                                & Get robot state                                                                                \\ \midrule
FSR                                                          & Force during hug                                                                            & \begin{tabular}[c]{@{}l@{}}Get force and \\ its location of exertion\\ during hug\end{tabular} \\ \midrule
\begin{tabular}[c]{@{}l@{}}Myo\\ armbands\end{tabular}       & \begin{tabular}[c]{@{}l@{}}Person's arm\\ acceleration\\ and orientation\end{tabular}       & \begin{tabular}[c]{@{}l@{}}Get arm movements\\ of the person during\\ hug\end{tabular}         \\ \midrule
Shoes                                                        & Foot pressure                                                                               & Get gait state                                                                                 \\ \midrule
\begin{tabular}[c]{@{}l@{}}Reflective\\ markers\end{tabular} & \begin{tabular}[c]{@{}l@{}}Position and \\ orientation\\ of robot and\\ person\end{tabular} & \begin{tabular}[c]{@{}l@{}}Annotate sensor data.\\ Ground truth.\end{tabular}                  \\ \bottomrule

\end{tabular}
\caption{Summary of features extracted from the devices/sensors and the hypothesis during designing the experiment. (FSR= Force Sensitive Resistor)}
\label{tab:feature_summary}
\end{table}


\section{Discussion}
\label{discussion}

We extracted and analyzed features from the data obtained from the robot movement and all the sensors. 
Using the mo-cap data, we recorded the position of the person and the robot in the room.
We used that position data to annotate start and end of hug and to annotate other sensor data into "before hug", "during hug" and "after hug" periods.
The features that indicate certain characteristics of hug activity are discussed below.

\begin{table}[ht]
\centering
\begin{tabular}{@{}lcc@{}}
\toprule
\multicolumn{2}{c}{\textbf{FSR Location}} & \textbf{\begin{tabular}[c]{@{}c@{}}Number of times\\ person touched (\%)\end{tabular}} \\ \midrule
                & upper forearm           & 58.07                                                                                  \\
Right           & lower forearm           & 50.99                                                                                  \\
                & wrist                   & 62.89                                                                                  \\\midrule
                & upper forearm           & 26.63                                                                                  \\
Left            & lower forearm           & 54.67                                                                                  \\
                & wrist                   & 45.33                                                                                  \\ \bottomrule
\end{tabular}
\caption{Summary of touches on robot's arms shows that the robot's upper arm (right) is more prominent in touching the person during the hug.
Total 353 hugs. (FSR= Force Sensitive Resistor)}
\label{tab:fsr_summary}
\end{table}

\begin{figure}[ht]
\centering
\includegraphics[scale=0.7]{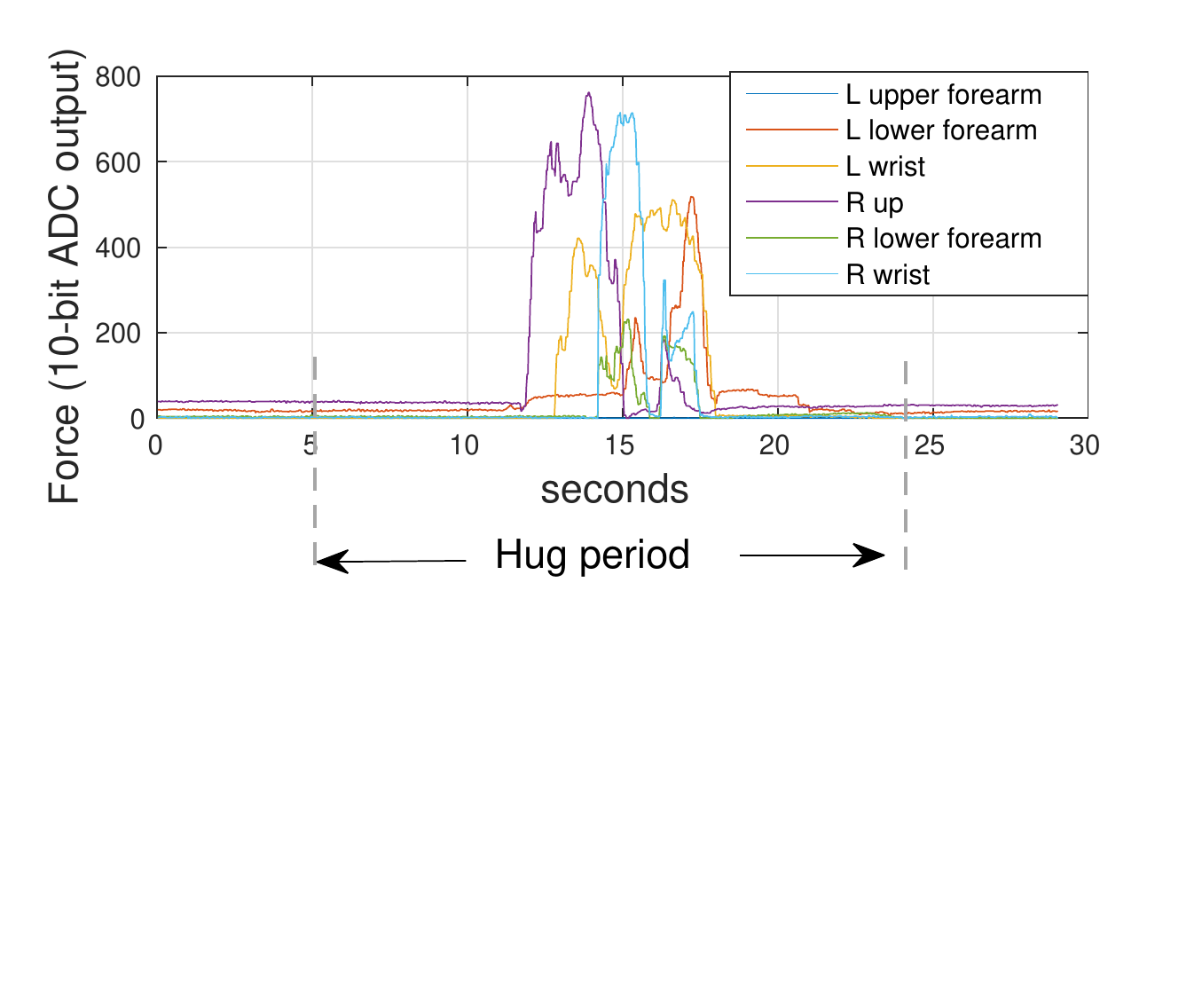}
\vspace{-38mm}
\caption{FSR values of a hug episode. Second 5-10: robot closing its arms for the hug. Second 10-18: robot and person grasp each other for the hug which results in force on the FSRs on robot's arms. Second 19-25: robot opening its arms to let the person go. (L= left, R= right)}
\label{fig:fsr_values}
\end{figure}

The FSR features indicate which parts of the robot arm touched the person during the hug.
They also indicate how tightly the robot was used to hug the person.
The author, who controlled the robot, moved the robot's arm after seeing how the person was approaching the robot for the hug.
The participants were not told how to hug the robot. 
In some hugs, people kept both their arms under the robot's arms, some times above the robot's arms and some times a cross-like arm position, seen in Figure \ref{fig:hug_screenshots}.
Table \ref{tab:fsr_summary} summarizes the number of times the person touched each FSR on robot's arm.
It can be seen that the robot's right arm was prominent in touching the person.
It should be noted that the robot's right arm was above its left arm in the initial pose, as seen in Figure \ref{fig:baxter_init}. 
This resulted in the right arm staying above the left in more than 90\% of the hugs.
A person's upper back is broader that lower back or waist.
So the robot's upper arm, right arm in most hugs, could lay flat on a person's upper back.
The robot arm is not as dexterous as a human arm to wrap around very well on the lower back region of a person. 
So the robot's lower arm, left arm in most hugs, hovered over the person's lower back, if it did not touch the person.
Figure \ref{fig:fsr_values} demonstrates that the FSR features are useful to determine physical contact during hug.
For future autonomous hugging task, we plan to implement a safety component in software that checks on the FSRs to exert safe forces during such close physical interaction task of hugging.

\begin{figure}[ht]
\centering
\includegraphics[scale=0.7]{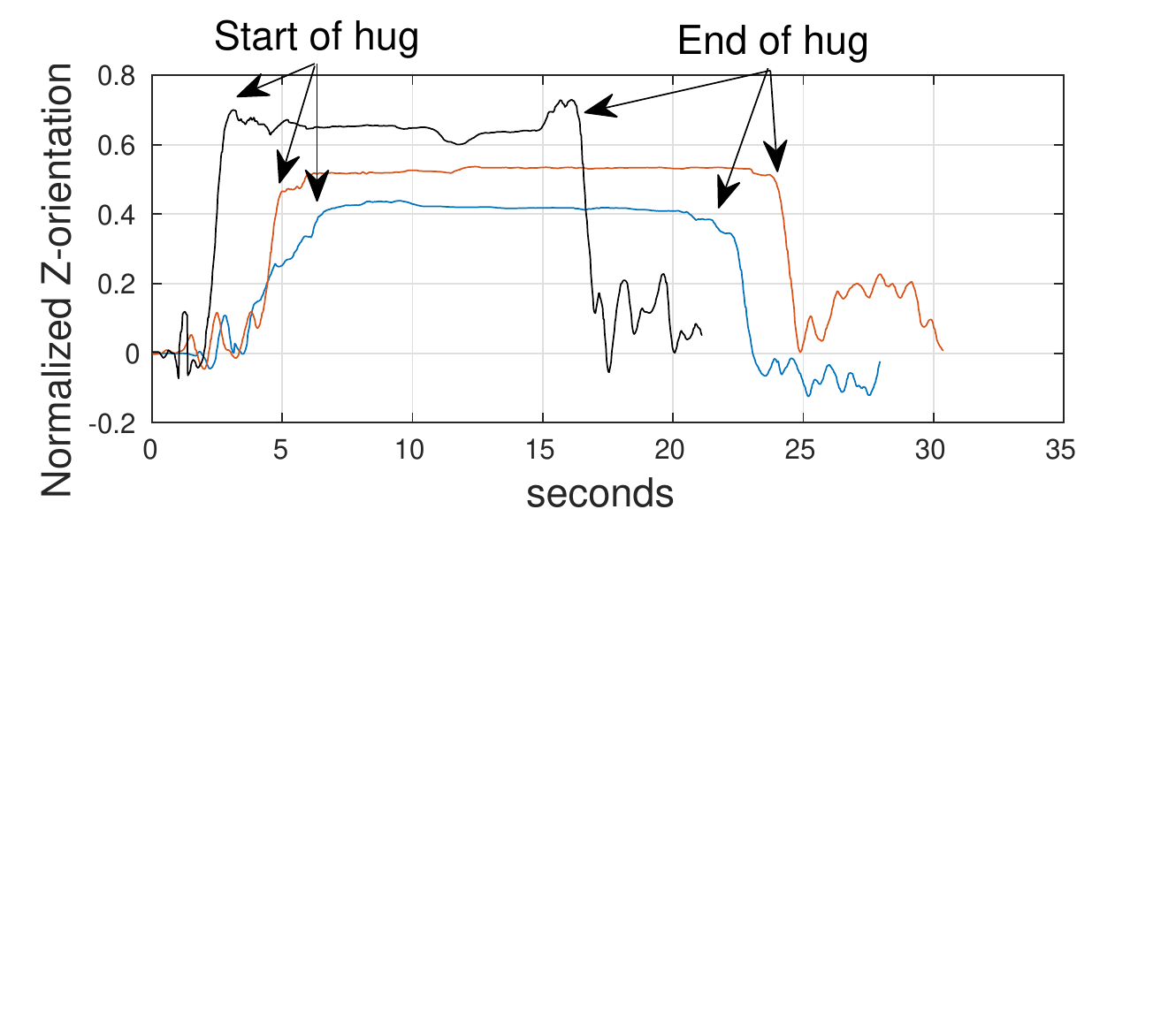}
\vspace{-43mm}
\caption{Orientation signals of left arm along Z-axis during hug episodes of 3 participants. Only 3 plots shown for clarity in figure. All hugs produced similar signal}
\label{fig:z_ori}
\end{figure}

From the Myo armbands, the orientation data is sent as quaternion.
Its w,x,y,z co-efficients are in [-1,1] range. 
Figure \ref{fig:z_ori} shows orientation of people's \textit{left} arm along Z-axis.
The z-orientation has a pattern that can indicate start and end of hug. 
As the person raises his arms to wrap around the robot, the z-orientation signal increases from zero and stays high till the person's arms are wrapped around the robot.
As seen in Figure \ref{fig:z_ori}, the z-orientation drops to nearly zero whenever the hug ends.
The z-orientation of the person's \textit{right} arm also produces signal of similar nature.
This feature is significant to identify that the person has started the hug and the robot can learn to react to it.

\begin{figure}[ht]
\centering

\includegraphics[scale=0.7]{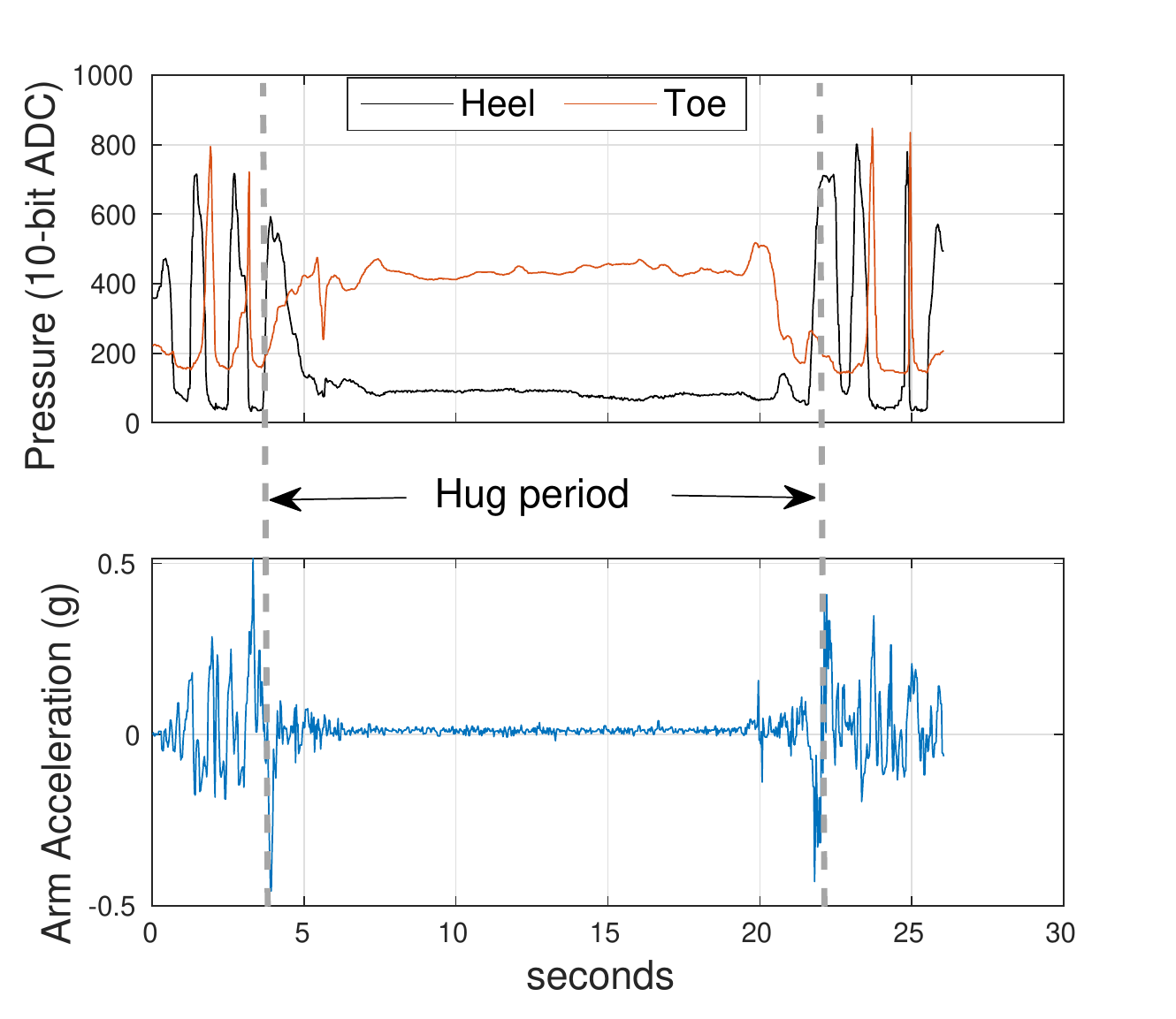}
\caption{Sensor signals that distinguish walking and stationary states of a person in this 26\textit{s} hug episode. Top: Right shoe's toe and heel pressure signals. Left shoe signals have similar nature. Bottom: Acceleration of right Myo armband.}
\label{fig:shoes_accel}
\end{figure}

Features that are useful in inferring gait of the person are body acceleration and foot pressure.
The top part of Figure \ref{fig:shoes_accel} shows heel and toe pressure signals of right foot of a person during a hug episode.
The person reached the robot in 4\textit{s} and began the hug which lasted till 23\textit{s} after which the person walked back to the starting position.
The shoes contain four pressure sensors but the signals from two of them, heel and toe, are enough to distinguish walking from standing. 
The addition of these two features creates a single and similar feature and will be used for feature reduction.
Left shoe produces similar signals.
The bottom part of Figure \ref{fig:shoes_accel} shows arm acceleration, derived from X-, Y-, Z-acceleration signals of right Myo armband.
The acceleration is nearly zero when the person is hugging the robot.
The arm swing during approach for hug and walking away from robot generates a distinct acceleration signal, as seen in Figure \ref{fig:shoes_accel}.

\begin{figure}[ht]
\centering
\hspace{-2mm}
\includegraphics[scale=0.68]{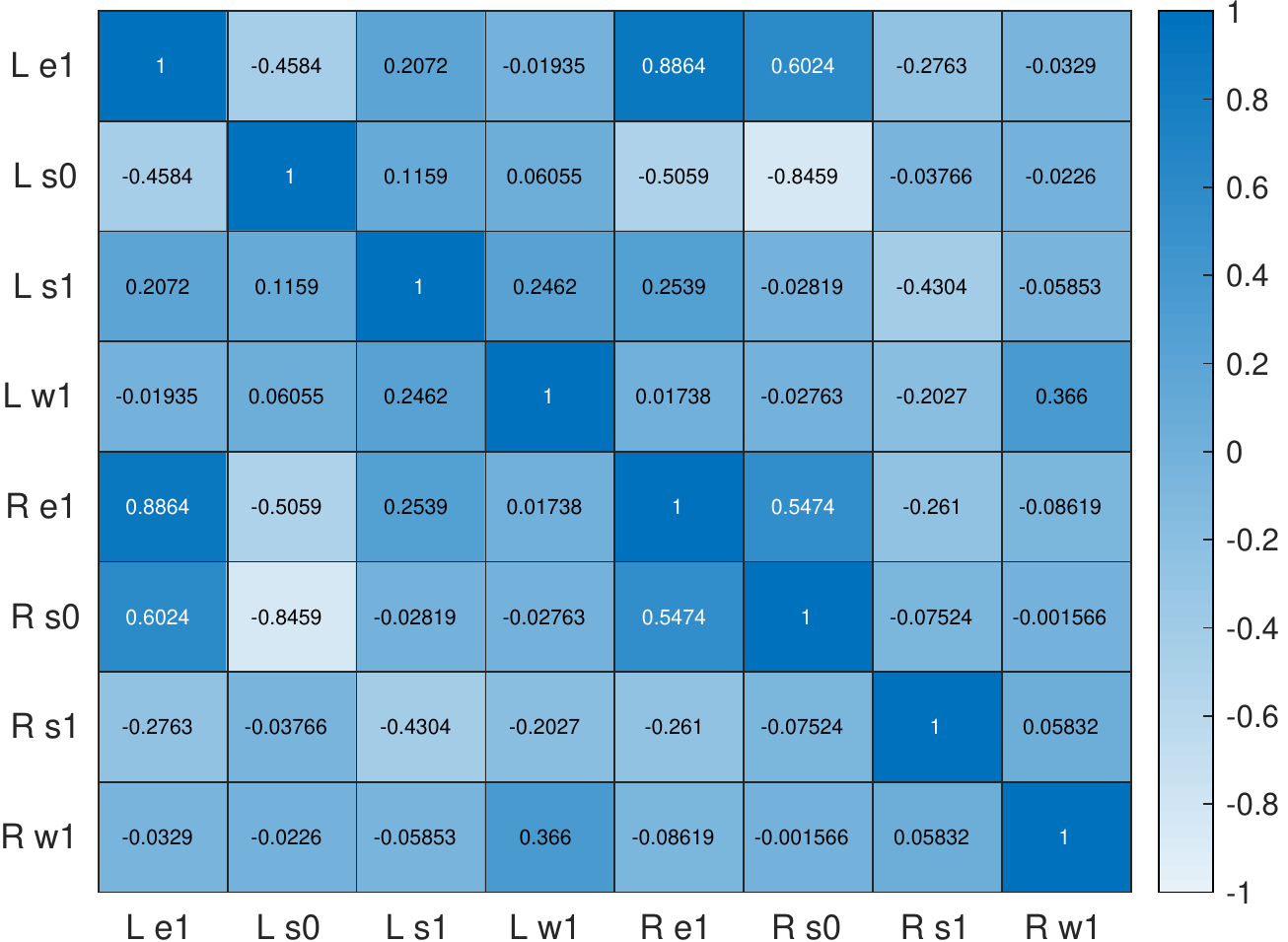}
\vspace{1mm}
\caption{Color-map of correlation between DOFs of the robot arms. (L =left, R =right) }
\vspace{-3mm}
\label{fig:corr_dof}
\end{figure}

The robot DOFs are labelled in Figure \ref{fig:joint_labels}.
All \textit{seven} DOFs of each of the robot's arms are controllable. 
However, for the hugging action, there are \textit{four} DOFs on each arm that were used the most: shoulder pitch(\textit{s1}) for adjusting to the height of the person, shoulder roll(\textit{s0}) and elbow pitch(\textit{e1}) for closing and opening arms for the hug and wrist pitch(\textit{w1}) for wrapping the robot arm around the person's lower back region.
The wrist link on the robot arm corresponds to the hand on a human arm.
The Pearson's correlation analysis, plotted as a color-map in Figure \ref{fig:corr_dof}, reveals correlation between seceral DOFs as expected.
The robot-operator opened and closed both the robot's arms simultaneously to make a human-like hugging action.
So the pairs of \textit{left e1} and \textit{right e1}, \textit{left s0} and \textit{right s0} have high correlation coefficient of 0.8864 and -0.8459 respectively.
In fact, the trajectories of \textit{s0} and \textit{e1} of both arms are well correlated with each other as depicted in the color-map.

\begin{figure}[ht]
\centering
\includegraphics[scale=0.6]{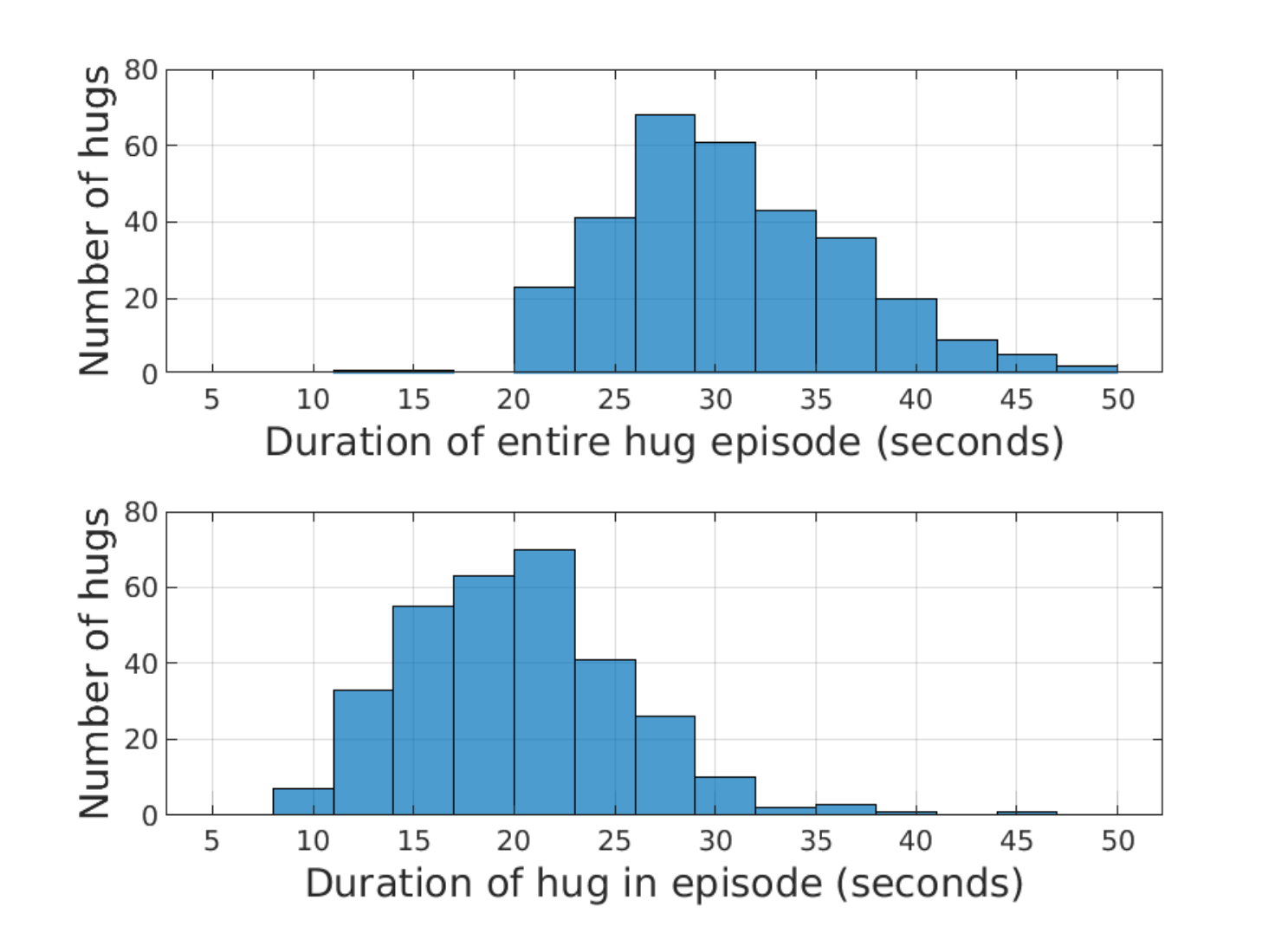}
\caption{Top: Histogram of duration of hug \textit{episodes}. Bottom: Histogram of duration of \textit{hug period} in the episode. }
\label{fig:duration_histogram}
\end{figure}

A hug episode consists of the person walking towards the robot, hugging the robot and walking back.
As mentioned in experiment procedure, the participants were given minimal instructions for hugging.
They took their own time for each of these tasks.
This resulted in variety in the collected dataset with respect to the duration of the episode and the actual hug period in the episode, as seen in the histograms in Figure \ref{fig:duration_histogram}.
The statistics of these durations is shown in Table \ref{tab:duration_stats}.
The average hug duration of \textit{20.11s} looks longer than average human-human hug because we operated the robot at a slow, safe speed of \textit{0.2rad/s}.

\begin{table}[ht]
\centering
\begin{tabular}{@{}lcccc@{}}
\toprule
\multicolumn{1}{c}{\textbf{}} & \textbf{Mean} & \textbf{Min} & \textbf{Max} & \textbf{Std Dev} \\ \midrule
Episode Duration (\textit{s})          & 30.72         & 13.82        & 57.62        & 6.21             \\
Hug Duration (\textit{s})              & 20.11         & 10.15        & 45.88        & 5.45             \\ \bottomrule
\end{tabular}
\caption{ Statistics of duration of episodes and hug period}
\label{tab:duration_stats}
\end{table}

The robot-operator sat behind the participant during the experiment to avoid obstructing the motion-capture space in the room and to have a clear sight of the person and the robot.
So he was not visible to the participant during participant's approach to hug. 
During the course of their respective hugging episodes, at least three participants asked questions like "Does the robot hug tightly?" and "What if the robot does not hug me?", which indicate that they forgot the fact that a person was controlling the robot and it was not autonomous. 
This suggests that the robot's hugging action looked human-like and that the participant had immersive hugging experience.

\section{Conclusions and Future Work}
\label{conclusion}
A completely autonomous robot is expected to learn when and how to engage in a collaboration, in order to assist a human partner to achieve a given task.
We performed hugging experiment between Baxter robot and \textit{33} participants to collect \textit{353} hugging samples and presented our methodology.
In our analysis of the features extracted from the robot and the wearable sensors, we demonstrate the significance of the dataset by showing trends and marking important features in the data that the robot can learn to react to the person approaching for a hug, start and end the hug in a safe, intuitive and affable manner that promotes collaboration between the two agents. 
This learning is specific to the Baxter robot used in the experiment because the output of the learning algorithm will be a set of joint angles meant for Baxter robot. 

The dataset presented in this work lays the foundation for future work in developing human-robot interaction algorithms for social scenarios.
This is a difficult task often complicated by the lack of appropriate data.
However, the dataset resulting from this work enables strategies such as learning from demonstration in multimodal scenarios~\cite{modelling}.
This approach is particularly appealing as it enables spatiotemporal filtering of salient interaction features in the face of nonlinear dynamics.
We will investigate such methods and analyze their ability to perform both intention inference and response generation as applied to hugging.


\bibliographystyle{aaai}
\bibliography{mybib}            

\begin{thebibliography}{}

\bibitem[\protect\citeauthoryear{Bethel \bgroup et al\mbox.\egroup
  }{2018}]{dog_robot}
Bethel, C.~L.; Henkel, Z.; Darrow, S.; and Baugus, K.
\newblock 2018.
\newblock Therabot-an adaptive therapeutic support robot.
\newblock In {\em 2018 World Symposium on Digital Intelligence for Systems and
  Machines (DISA)},  23--30.
\newblock IEEE.

\bibitem[\protect\citeauthoryear{Block and Kuchenbecker}{2019}]{soft_warm}
Block, A.~E., and Kuchenbecker, K.~J.
\newblock 2019.
\newblock Softness, warmth, and responsiveness improve robot hugs.
\newblock {\em International Journal of Social Robotics} 11(1):49--64.

\bibitem[\protect\citeauthoryear{Campbell, Stepputtis, and
  Amor}{2019}]{modelling}
Campbell, J.; Stepputtis, S.; and Amor, H.~B.
\newblock 2019.
\newblock Probabilistic multimodal modeling for human-robot interaction tasks.
\newblock {\em Robotics: Science and Systems XV}.

\bibitem[\protect\citeauthoryear{Cooney \bgroup et al\mbox.\egroup
  }{2010}]{playful_robot}
Cooney, M.~D.; Becker-Asano, C.; Kanda, T.; Alissandrakis, A.; and Ishiguro, H.
\newblock 2010.
\newblock Full-body gesture recognition using inertial sensors for playful
  interaction with small humanoid robot.
\newblock In {\em 2010 IEEE/RSJ International Conference on Intelligent Robots
  and Systems},  2276--2282.
\newblock IEEE.

\bibitem[\protect\citeauthoryear{Forsell and
  {\AA}str{\"o}m}{2012}]{hug_meaning}
Forsell, L.~M., and {\AA}str{\"o}m, J.~A.
\newblock 2012.
\newblock Meanings of hugging: From greeting behavior to touching implications.
\newblock {\em Comprehensive Psychology} 1:02--17.

\bibitem[\protect\citeauthoryear{Hirokawa and Suzuki}{2018}]{hug_toy}
Hirokawa, E., and Suzuki, K.
\newblock 2018.
\newblock Design of a huggable social robot with affective expressions using
  projected images.
\newblock {\em Applied Sciences} 8(11):2298.

\bibitem[\protect\citeauthoryear{Jeong \bgroup et al\mbox.\egroup
  }{2015}]{hug_teddy}
Jeong, S.; Santos, K.~D.; Graca, S.; O'Connell, B.; Anderson, L.; Stenquist,
  N.; Fitzpatrick, K.; Goodenough, H.; Logan, D.; Weinstock, P.; et~al.
\newblock 2015.
\newblock Designing a socially assistive robot for pediatric care.
\newblock In {\em Proceedings of the 14th international conference on
  interaction design and children},  387--390.
\newblock ACM.

\bibitem[\protect\citeauthoryear{Jindai \bgroup et al\mbox.\egroup
  }{2017}]{hug_model}
Jindai, M.; Ota, S.; Yasuda, T.; Sasaki, T.; and Sejima, Y.
\newblock 2017.
\newblock Development of a hug request motion model during active approach to
  human.
\newblock In {\em 2017 IEEE International Conference on Systems, Man, and
  Cybernetics (SMC)},  612--617.
\newblock IEEE.

\bibitem[\protect\citeauthoryear{Kim \bgroup et al\mbox.\egroup
  }{2016}]{child_hug}
Kim, J.; Alspach, A.; Leite, I.; and Yamane, K.
\newblock 2016.
\newblock Study of children's hugging for interactive robot design.
\newblock In {\em 2016 25th IEEE International Symposium on Robot and Human
  Interactive Communication (RO-MAN)},  557--561.
\newblock IEEE.

\bibitem[\protect\citeauthoryear{Lee \bgroup et al\mbox.\egroup
  }{2014}]{cat_robot}
Lee, B.~H.; Jang, J.-y.; Mun, K.-h.; Kwon, J.~Y.; and Jung, J.~S.
\newblock 2014.
\newblock Development of therapeutic expression for a cat robot in the
  treatment of autism spectrum disorders.
\newblock In {\em 2014 11th International Conference on Informatics in Control,
  Automation and Robotics (ICINCO)}, volume~2,  640--647.
\newblock IEEE.

\bibitem[\protect\citeauthoryear{Mitchell}{1999}]{ML}
Mitchell, T.~M.
\newblock 1999.
\newblock Machine learning and data mining.
\newblock {\em Communications of the ACM} 42(11).

\bibitem[\protect\citeauthoryear{Miyashita and
  Ishiguro}{2004}]{involuntary_robot}
Miyashita, T., and Ishiguro, H.
\newblock 2004.
\newblock Human-like natural behavior generation based on involuntary motions
  for humanoid robots.
\newblock {\em Robotics and Autonomous Systems} 48(4):203--212.

\bibitem[\protect\citeauthoryear{Myo-Support}{2018}]{Myo_specs}
Myo-Support.
\newblock 2018.
\newblock Myo gesture control armband tech specs.
\newblock
  \url{https://support.getmyo.com/hc/en-us/articles/202648103-Myo-Gesture-Control-Armband-tech-specs}.
\newblock Accessed June 14, 2019.

\bibitem[\protect\citeauthoryear{Rethink-Robotics}{2015}]{Baxter_specs}
Rethink-Robotics.
\newblock 2015.
\newblock Baxter hardware specifications.
\newblock \url{http://sdk.rethinkrobotics.com/wiki/Hardware_Specifications}.
\newblock Accessed June 17, 2019.

\bibitem[\protect\citeauthoryear{Robinson \bgroup et al\mbox.\egroup
  }{2013}]{companion_robot}
Robinson, H.; MacDonald, B.; Kerse, N.; and Broadbent, E.
\newblock 2013.
\newblock The psychosocial effects of a companion robot: a randomized
  controlled trial.
\newblock {\em Journal of the American Medical Directors Association}
  14(9):661--667.

\bibitem[\protect\citeauthoryear{Shiomi \bgroup et al\mbox.\egroup
  }{2017a}]{robot_touch}
Shiomi, M.; Nakagawa, K.; Shinozawa, K.; Matsumura, R.; Ishiguro, H.; and
  Hagita, N.
\newblock 2017a.
\newblock Does a robot’s touch encourage human effort?
\newblock {\em International Journal of Social Robotics} 9(1):5--15.

\bibitem[\protect\citeauthoryear{Shiomi \bgroup et al\mbox.\egroup
  }{2017b}]{prosocial}
Shiomi, M.; Nakata, A.; Kanbara, M.; and Hagita, N.
\newblock 2017b.
\newblock A hug from a robot encourages prosocial behavior.
\newblock In {\em 2017 26th IEEE international symposium on robot and human
  interactive communication (RO-MAN)},  418--423.
\newblock IEEE.

\bibitem[\protect\citeauthoryear{Stiehl \bgroup et al\mbox.\egroup
  }{2009}]{hug_platform_teddy}
Stiehl, W.~D.; Lee, J.~K.; Breazeal, C.; Nalin, M.; Morandi, A.; and Sanna, A.
\newblock 2009.
\newblock The huggable: a platform for research in robotic companions for
  pediatric care.
\newblock In {\em Proceedings of the 8th International Conference on
  interaction Design and Children},  317--320.
\newblock ACM.

\bibitem[\protect\citeauthoryear{Yaseen and Lohan}{2018}]{robot_trust}
Yaseen, A., and Lohan, K.
\newblock 2018.
\newblock Playing pairs with pepper.
\newblock {\em arXiv preprint arXiv:1810.07593}.

\end{thebibliography}

\end{document}